\documentclass[lettersize,journal]{IEEEtran}
\usepackage{amsmath,amsfonts}

\usepackage{algorithm}
\usepackage{array}
\usepackage[caption=false,font=normalsize,labelfont=sf,textfont=sf]{subfig}
\usepackage{textcomp}
\usepackage{stfloats}
\usepackage{url}
\usepackage{verbatim}
\usepackage{graphicx}
\usepackage{cite}
\usepackage{graphicx}
\usepackage{amsmath}
\usepackage{amsthm}
\usepackage{enumitem}
\usepackage{adjustbox}
\newtheorem{definition}{Definition}
\newtheorem{proposition}{Proposition}
\newtheorem{lemma}{Lemma}

\newtheorem{assumption}{Assumption}
\usepackage{hyperref}       
\usepackage{url}            
\usepackage{booktabs}       
\usepackage{amsfonts}       
\usepackage{nicefrac}       
\usepackage{microtype}      
\usepackage{xcolor}         
\newcommand{\shortname}{\texttt{Waffle} }

\usepackage{algorithm}
\usepackage{algpseudocode}
\usepackage{multirow}

\begin{document}

\title{Wavelet Scattering Transform and Fourier Representation for Offline Detection of Malicious Clients in Federated Learning}

\author{
Alessandro Licciardi*\thanks{
A.L. is affiliated to the Department of Mathematical Sciences, Politecnico di Torino, Turin, Italy, and to the
Istituto Nazionale di Fisica Nucleare, Sezione di Torino, Turin, Italy }
\and
Davide Leo$^*$\thanks{
D.L. works for Tynk S.R.L., Via Viberti 4, Turin, Italy
}
\and
Davide Carbone \thanks{
 D.C. is affiliated to Laboratoire de Physique de l’Ecole Normale Supérieure, ENS Université PSL; and to the CNRS, Sorbonne Université, Université de Paris, Paris, France
}
\thanks{Correspondance to \texttt{alessandro.licciardi@polito.it}}\thanks{* Denotes equal contributions}}
\maketitle

\thispagestyle{plain}
\pagestyle{plain}

\renewcommand{\thefootnote}{\fnsymbol{footnote}} 


\maketitle

\begin{abstract}
Federated Learning (FL) enables the training of machine learning models across decentralized clients while preserving data privacy. However, the presence of anomalous or corrupted clients—such as those with faulty sensors or non-representative data distributions—can significantly degrade model performance. Detecting such clients without accessing raw data remains a key challenge. We propose \shortname  (\textbf{Wa}velet and \textbf{F}ourier representations for \textbf{F}ederated \textbf{Le}arning) a detection algorithm  that labels malicious clients {\it before training}, using locally computed compressed representations derived from either the Wavelet Scattering Transform (WST) or the Fourier Transform. Both approaches provide low-dimensional, task-agnostic embeddings suitable for unsupervised client separation. A lightweight detector, trained on a distillated public dataset, performs the labeling with minimal communication and computational overhead. While both transforms enable effective detection, WST offers theoretical advantages, such as non-invertibility and stability to local deformations, that make it particularly well-suited to federated scenarios. Experiments on benchmark datasets show that our method improves detection accuracy and downstream classification performance compared to existing FL anomaly detection algorithms, validating its effectiveness as a pre-training alternative to online detection strategies. Source code for the paper is publicly available at \url{https://github.com/davedleo/Waffle}.

\end{abstract}

\newcommand{\var}{\mathbb{V}ar}

\section{Introduction}
Federated Learning (FL) is a key paradigm for enabling decentralized intelligence in large-scale Internet of Things (IoT) systems, allowing numerous devices to train a shared model without exposing raw data \cite{mcmahan2017communication, bonawitz2019towards}. This approach is vital for privacy-sensitive applications in domains like smart cities, industrial automation, and connected healthcare \cite{antunes2022federated, long2020federated}. However, the success of FL in the IoT is threatened by two intertwined challenges: vast data heterogeneity from diverse devices and the system's vulnerability to malicious or faulty clients \cite{li2020federated, yin2018byzantine, cao2021fltrust}.

Consider a smart factory deployment where an Industrial IoT network monitors equipment health \cite{dzaferagic2021fault}. Some sensors may become miscalibrated due to environmental stress, suffer physical damage, or be deliberately compromised, injecting anomalous data into the system. Such data poisoning attacks can degrade the global model's performance, leading to incorrect predictions about machine failure. Detecting and isolating these compromised devices before they participate in training is crucial for maintaining network integrity and operational safety \cite{agrawal2022federated}. While defenses like robust aggregation \cite{blanchard2017machine,yin2018byzantine} can mitigate the impact of malicious model updates, they are often less effective against subtle data-level perturbations and typically assume a majority of clients are benign—a guarantee that may not hold in a widely distributed and physically accessible IoT network. Furthermore, online detectors that monitor clients during training can introduce significant communication and computational overhead, which is untenable for resource-constrained IoT devices, and may only flag threats after the damage has begun.

In this work, we propose \shortname (\textbf{Wa}velet and \textbf{F}ourier representations for \textbf{F}ederated \textbf{Le}arning), a lightweight, offline detector designed to identify and exclude clients with malicious data before FL training begins. Its privacy-preserving, pre-training detection is ideal for scalable IoT deployments, as it minimizes computational overhead on both the server and the edge devices. \shortname trains a classifier on stable spectral features—extracted via the Fourier Transform (FT) and Wavelet Scattering Transform (WST) \cite{mallat2012group}—which provide robust representations of client data distributions. The detection is performed using a model pre-trained on a public dataset, ensuring efficiency and privacy. Clients only need to compute low-dimensional spectral statistics and send a secure, non-invertible summary to the server for classification. Unlike existing methods, \shortname remains effective even when malicious clients form a large majority. Our experiments demonstrate its high efficacy in diverse settings, including under challenging non-Gaussian data attacks, and we showcase its versatility with a proof-of-concept on a Natural Language Processing (NLP) task.

The structure of the paper is the following: Section \ref{sec:theoretical_framework} defines the FL setting, the data attacks considered, and the spectral representations (FT and WST). Section \ref{sec:detector} details \shortname's training and detection. Theoretical guarantees are provided in Section \ref{sec:theoretical_guarantees}, showing the benefits of removing malicious clients. Section \ref{sec:experiments} reports experimental results validating \shortname on benchmark datasets.

\subsection*{Related Work and Contributions.}
\textbf{Malicious Client Detection in FL}
Detection-based approaches classify clients as benign or malicious based on anomalies in their updates or data distribution \cite{fung2018mitigating}. \texttt{FLDetector}\cite{zhang2022fldetector} identifies malicious clients by analyzing the consistency of their updates over time—benign updates follow predictable patterns, while malicious ones are erratic. \texttt{MuDHog}\cite{gupta2022long} leverages historical update trajectories with model-agnostic meta-learning to detect temporal inconsistencies, though it assumes long-term client participation, which is unrealistic in cross-device settings. \texttt{VAE}~\cite{li2020learning} uses a variational autoencoder to model the benign update distribution and flags deviations, assuming malicious clients are rare and the VAE is well-trained. These methods rely on multi-round update access, limiting early-stage applicability, and focus on gradients or parameters, making them vulnerable to indirect attacks.

\textbf{Robust Aggregation in FL.} Robust aggregation methods aim to mitigate the influence of malicious clients without explicitly identifying them \cite{guerraoui2018hidden,yin2018byzantine}. \texttt{KRUM}\cite{blanchard2017machine} selects the most central update in $\ell_2$ distance, but requires fewer than half of the clients to be malicious. \texttt{TrimmedMean}\cite{yin2018byzantine} discards extreme values per coordinate, improving robustness to outliers, though it overlooks dependencies across dimensions. \texttt{FLTrust}\cite{cao2021fltrust} uses a trusted server-side dataset to normalize and rescale client updates, enhancing robustness but breaking strict decentralization. Secure aggregation protocols like \texttt{RFLPA}\cite{mai2024rflpa} and \texttt{RoFL}~\cite{lycklama2023rofl} ensure client privacy via cryptographic techniques, but do not address adversarial robustness. These approaches, unlike detection-based ones, do not label clients, limiting their use when malicious participants must be explicitly excluded.

\textbf{Spectral Analysis and Frequency-based Defenses.} Spectral methods aim to identify or mitigate malicious behavior by analyzing updates in the frequency domain \cite{wang2024revisiting, chan2000fabric, tao2019hyperspectral}. \texttt{FreqFeD}\cite{fereidooni2023freqfed} applies the Discrete Fourier Transform to client updates, filtering high-frequency components assumed to contain adversarial noise, though this may remove relevant information under data heterogeneity. \texttt{FedSSP}\cite{ChenT24} targets backdoor attacks by smoothing and pruning suspicious spectral patterns in model weights, but depends on specific architectures and requires access to full model parameters. Unlike these methods, our approach extracts frequency-based embeddings directly from client-side data before training, enabling model-agnostic detection.

Our {\it main contributions} are summarized as follows:
\begin{itemize}[leftmargin=*, itemsep=0cm,topsep=0em]
\item We propose \shortname, a novel offline detector for identifying clients with data attacks, introducing the use of WST for anomaly detection in FL.
\item We provide a theoretical framework motivating WST and FT as robust data representations, and mathematically demonstrate that removing malicious clients improves global model estimation.
\item We present experiments on benchmark datasets showing that \shortname significantly improves model performance and robustness compared to training with contaminated data or using only robust aggregation.
\end{itemize}
\section{Theoretical Framework}\label{sec:theoretical_framework}
In this section, we introduce the mathematical framework that provides the foundation for our algorithm. Section \ref{sec:problem_form} presents the Federated Learning (FL) setting and defines the class of attacks considered on clients' data. Section \ref{sec:wst_ft} introduces the WST and the Fourier Transform FT, recalling their basic properties that are relevant for anomaly detection.
\subsection{Problem Formulation}\label{sec:problem_form}
Consider a standard FL scenario \cite{mcmahan2017communication} with $K \in \mathbb{N}$ clients and a central server. Each client $k$ possesses $n_k$ data samples $(x^i_k, y^i_k)_{i = 1}^{n_k} \sim \mathcal{D}_k$ supported in $\mathcal{X} \times \mathcal{Y}$. The objective of FL is to learn a shared global model $\theta$ that generalizes across all clients, by solving the following optimization problem:
\begin{equation}\label{eq:FL_opt}
\theta^* \in \arg \min_{\theta \in \Theta} \dfrac{1}{N} \sum_{k = 1}^K n_k \mathcal{L}_k(\theta)
\end{equation}
where $\Theta$ denotes the model's parameter space, $N = \sum_{k=1}^K n_k$ is the total number of data samples, and $\mathcal{L}_k$ represents the empirical loss function for client $k$ with respect to its local data distribution $\mathcal{D}_k$. In each communication round $t \in \{1,\dots, T\}$, a subset of clients $\mathcal{P}_t$ is randomly selected to participate in training. Each participating client $k \in \mathcal{P}_t$ performs $S \in \mathbb{N}$ local iterations of a stochastic optimizer. Subsequently, clients send their updated parameters to the server, which aggregates these updates to derive a new global model.\\
A critical challenge in realistic FL deployments is the {\it non-i.i.d.} nature of client data, which can hinder the convergence and performance of the global model. In this work, we specifically address non-i.i.d. settings where the data distribution discrepancies are caused by malicious clients perturbing their original data samples. This differs from typical attack detection scenarios focusing on model poisoning during training.

\begin{figure*}[!t]
    \centering
    \includegraphics[width=0.49\textwidth]{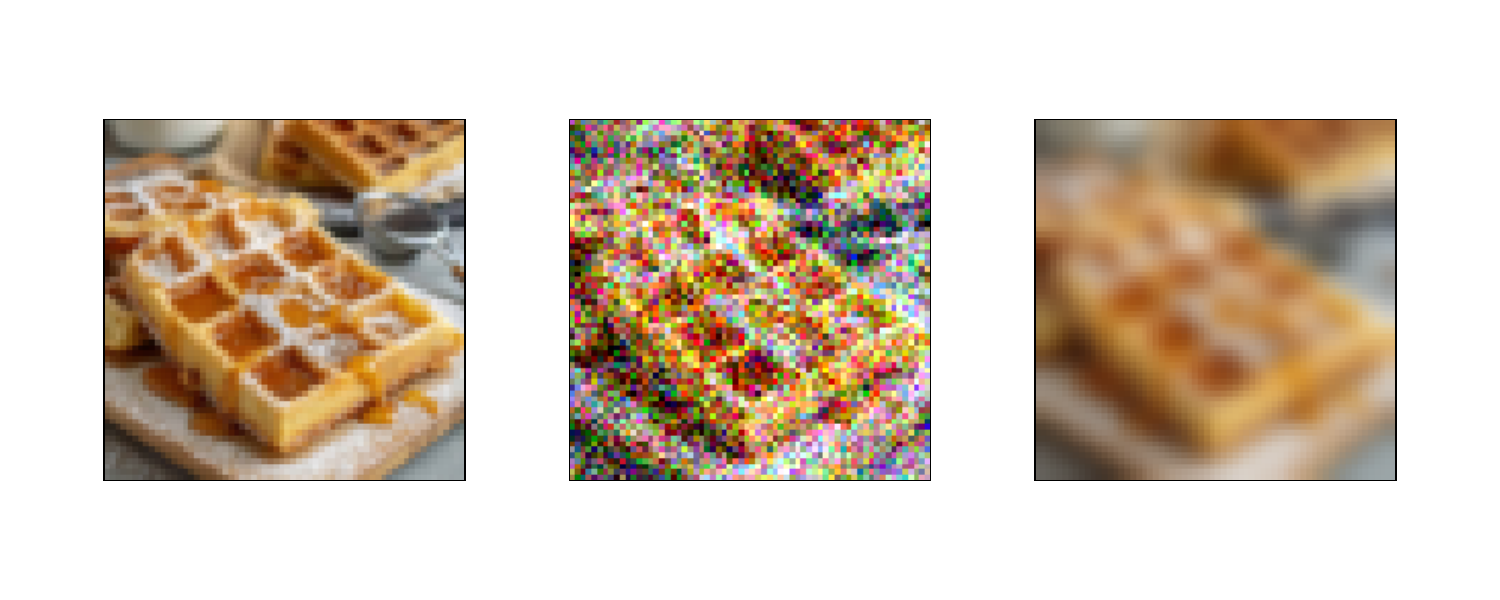}
    \hfill
    \includegraphics[width=0.49\textwidth]{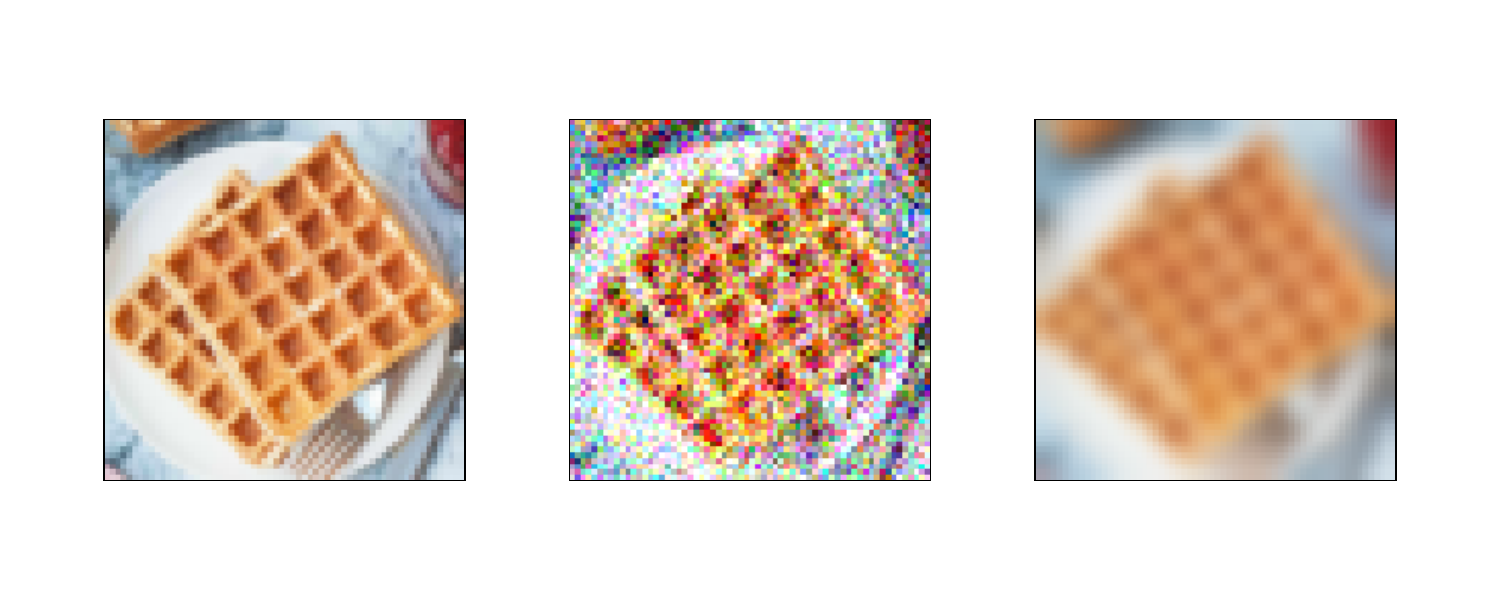}
    \caption{\small Examples of attacked data. Two images downloaded from \href{https://chefjar.com/wp-content/uploads/2024/12/belgian-waffle-recipe-1-1-1000x1477.jpg}{link1} and
    \href{https://www.google.com/imgres?imgurl=https://lh3.googleusercontent.com/C8p-ppuCYDwLOcZC-5SXWrmCywhDWV21vh6ri4XIHA_ZiHtJ8vC2WM4FejFA5WGB3iCs&tbnid=iM2fyTR1KCsKWM&vet=1&imgrefurl=https://apptopia.com/google-play/app/ru.rukart.VafliVafeln/about&docid=8aPyeFUJfaAqcM&w=512&h=512&hl=en-IT&source=sh/x/im/m1/1&kgs=f2519ed18262765a}{link2}.
    For each image: \textit{left}: clean client, \textit{center}: noisy attack with magnitude $\sigma = 0.2$, \textit{right}: blur attack with spread $\beta = 11$.}
    \label{fig:attack_Examples}
\end{figure*}

\paragraph{Type of Attacks}
We define two types of feature-level attacks that our algorithms aim to address: noisy and blur  attackers. Examples of the effect of these attacks are displayed in Figure \ref{fig:attack_Examples}. This focus is motivated by the fact that noise and blur are common consequences of real-world faults \cite{sharma2010sensor,peng2021motion} --such as sensor degradation, miscalibration, or environmental interference -- that can subtly compromise data quality and model performance without exhibiting overtly malicious behavior.
\begin{definition}\label{def:noisy_Attack}
Let $k \in [K]$ and $\sigma_k > 0$. Client $k$ is a \textbf{noisy attacker} if its data samples are perturbed as $ \tilde{x}_k^i = x_k^i + \sigma_k\epsilon_k^i$, where $x_k^i$ is the clean sample, and, $(\epsilon_k^i)_{i = 1}^{n_k}$ is a family of independent Wiener processes supported in $\mathcal{X}$. 
\end{definition}

Let us observe that the severity of the attack is determined by the magnitude of $\sigma_k$. Smaller values of $\sigma_k$ might represent natural noise inherent in data collection or random transformations, requiring careful consideration of what constitutes a 'malicious' level of perturbation.\\
Another feature-wise attack we formally define is the \textbf{blur attacker}. This attack is particularly relevant for image or signal data where $x_k^i$ can be treated as a function over $\mathcal{X}$.

\begin{definition}\label{def:blur_attack}
Let $k \in [K]$ and $\beta_k > 0$. Client $k$ is a \textbf{blurred attacker} if it provides samples perturbed according to a convolution operation:
\begin{equation}\label{eq:blur_conv}
\tilde{x}_k^i = x_k^i \star \zeta_k = \int_{\mathcal{X}} x_k^i(u') \zeta_k(u - u') du' \quad i = 1,\dots,n_k
\end{equation}
where $\star$ denotes the convolution operation. Typically, $\zeta_k$ is a smooth kernel, and the parameter $\beta_k$ controls its spread or blur radius. 
\end{definition}
A common choice for the kernel $\zeta_k$ falls on Gaussian kernels, and the scalar $\beta_k$ has a role of controlling the spread of the kernel. Similarly to noisy attacks, in blur attacks the magnitude of the perturbation is controlled by the parameter $\beta_k$, the larger it is the higher it perturbs the data.

\subsection{Representation Operators: Wavelet Scattering Transform and Fourier Transform}\label{sec:wst_ft}
In this section we recall the notion of a representation operator $\Phi$, which maps a signal $x$ (e.g., an image or a time-series) onto a transformed space. This transformation induces a metric $d(x,x') = \|\Phi[x] - \Phi[x']\|$ in the new space \cite{bruna2013invariant}. The core idea is that an effective representation operator $\Phi$ should possess properties instrumental for accurately detecting and differentiating between data samples. Specifically, for the purpose of identifying perturbed data, $\Phi$ should be able to separate distinct data characteristics while exhibiting robustness to common variations like slight translations or small, non-malicious perturbations. We propose two variants for the representation layer of our detection algorithm: one based on the Fourier Transform (FT) and the other on the Wavelet Scattering Transform (WST) \cite{mallat2012group, bruna2013invariant}. The Fourier Transform is by far the most widely used tool for spectral analysis in signal processing and data science due to its simplicity and interpretability. However, it has been surprisingly underutilized in the context of Federated Learning (FL). We therefore include it as an internal baseline in our study, allowing us to contrast its performance against the more structured and hierarchical Wavelet Scattering Transform.
\paragraph{Fourier Representation} 
We first formally define the Fourier Transform.
\begin{definition}
    Let $x \in L^1(\mathcal{X}, du)$, the \textbf{Fourier Transform} of $x$, denoted by $\mathcal{F}[x]$ is a complex valued function defined as
    \begin{equation}\label{eq:FT}
        \mathcal{F}[x](\omega)= \int_\mathcal{X} x(u) e^{-2\pi i (u \cdot \omega)} du
    \end{equation}
\end{definition}
FT can be efficiently computed using the FFT algorithm \cite{cooley1965algorithm}. Beyond its computational efficiency, the Fourier Transform offers several critical advantages for feature extraction, particularly in the context of analyzing data perturbations. As a linear operator ($\mathcal{F}[ax+bx'] = a\mathcal{F}[x] + b\mathcal{F}[x']$ for scalars $a,b$ and integrable signals $x,x'$), the FT maps additive perturbations directly to additive components in the frequency domain. For instance, in the case of a \textit{noisy attacker} where $\tilde{x} = x + \epsilon$, we have $\mathcal{F}[\tilde{x}] = \mathcal{F}[x] + \mathcal{F}[\epsilon]$. This linearity simplifies the analysis of such perturbations. Moreover, the convolution theorem \cite{bracewell1989fourier} states that convolution in the spatial domain corresponds to point-wise multiplication in the frequency domain ($\mathcal{F}[x \star \delta] = \mathcal{F}[x] \cdot \mathcal{F}[\delta]$) . This property is highly advantageous for detecting \textit{blur attacker} perturbations, which are defined as convolutions. By examining the frequency spectrum, different types of data manipulations, like blurring (attenuating high frequencies) or specific noise patterns, reveal distinct signatures. However, FT is an invertible operator: on one side it preserves all information present in the original signal, on the other hand it is possible to reconstruct the original data from the FT.   
\paragraph{Wavelet Scattering Transform.} WST is a non-linear operator that, alternatively to Fourier based representation, has been designed to be stable to additive perturbations, locally translation invariant and to small continuous deformation. Moreover, the fact that WST is not invertible makes it particularly attractive for privacy-enhancing applications in FL, as reconstructing the original input data from the scattering coefficients is a challenging task. Following the construction in \cite{mallat2012group,bruna2013invariant} we define the WST and discuss its most relevant properties.\\
Let $\psi(u) \in L^2(\mathcal{X}, du)$ be a function referred to as the \textbf{mother wavelet}, and let $\{a^j\}_{j \in \mathbb{Z}}$ be a family of scale factors defined with respect to a fixed scalar $a > 1$. Let $r \in G$ denote a discrete rotation, where $G$ is the group of discrete rotations acting on the domain $\mathcal{X}$. The $j$-th \textbf{wavelet function} is then defined as $\psi_j(u) = a^{-dj} \psi(a^{-j} r^{-1} u)$. For a fixed maximal depth $J \in \mathbb{Z}$, we define the set of admissible scale-rotation operators as $\Lambda_J = \{\lambda = a^j r : |\lambda| = a^j < 2^J\}.$ In most implementations, Morlet wavelets are employed as the mother wavelet, and the scale factor is typically chosen as $a = 2^{1/Q}$ for some $Q \in \mathbb{N}$ \cite{andreux2020kymatio}. \\
To streamline notation, following \cite{mallat2012group}, we introduce the \textbf{propagator operator}, which acts on a signal $x \in L^1(\mathcal{X})$ by cascading modulus and convolution operations. Given a path of scale-rotation operators $p = (\lambda_1, \lambda_2)$, the propagator applied to $x$ is defined as:
\[
U[p]x = \left|\,|x \star \psi_{\lambda_1}| \star \psi_{\lambda_2}\,\right|.
\]
The definition of the WST naturally follows.

\begin{definition}\label{def:wst}
Let $p = (\lambda_1, \dots, \lambda_m) \subset \Lambda_J$ be a path of length $m$. For any signal $x \in L^1(\mathcal{X})$, the WST along $p$ is defined as:
\begin{equation}\label{eq:wst_def}
    S_J[p] x = U[p]x \star \phi_J,
\end{equation}
where $\phi_J$ is a low-pass filter rescaled to recover low-frequency content.
\end{definition}

The WST representation shares structural similarities with convolutional neural networks (CNNs), with the key distinction that the wavelet filters are fixed rather than learned. The WST defines a norm with properties desirable for detection and classification. Notably, the operator is \textbf{non-expansive}: for any $x, x' \in L^2(\mathcal{X}, du)$, the following inequality holds:
\begin{equation}\label{eq_nonexpansive}
    \|S_J[p] x - S_J[p] x'\| \leq \|x - x'\|.
\end{equation}
This implies that small, non-adversarial perturbations do not substantially affect the representation.

Additionally, WST is \textbf{translation invariant} in the limit: for a translated signal $x_c(u) = x(u - c)$ with $c \in \mathcal{X}$, we have
\[
\lim_{J \to \infty} \|S_J[p] x - S_J[p] x_c\| = 0.
\]
Finally, the WST is \textbf{Lipschitz continuous} with respect to small \( C^2 \)-diffeomorphisms. That is, if a signal $x$ undergoes a smooth deformation with small norm, the resulting change in the WST representation remains bounded.

\section{Malicious Client Detector: \shortname}\label{sec:detector}
This section details the architecture and training of our server-side detector, \shortname (\textbf{Wa}velet and \textbf{F}ourier representations for \textbf{F}ederated \textbf{Le}arning), designed to identify clients contributing potentially harmful updates based on their data characteristics. \shortname is a parametric classification model, trained offline on a generated auxiliary dataset $\mathcal{D}^{\text{aux}}$ to distinguish between benign and malicious clients. It operates by analyzing aggregated, non-privacy-leaking spectral embeddings of client data distributions.

\subsection{Offline Detector Training}
The training of the \shortname\ detector is conducted entirely offline, prior to the federated learning process. This approach offers several advantages: it avoids interfering with live FL rounds, allows for controlled generation of diverse malicious scenarios, and ensures the detector is fully trained and ready when FL begins. Coherently with common practices in FL frameworks utilizing auxiliary data \cite{wang2018dataset}, the server has access to a representative auxiliary dataset $\mathcal{D}^{\text{aux}}$. Algorithm \ref{alg:waffle_summary} summarizes the procedure.\\
The offline training proceeds over $E$ epochs. In each epoch $e \in \{1,\dots,E\}$, the server simulates a new FL round by generating a set of $\tilde{K}$ fictitious clients with synthetic data and associated ground-truth labels (benign or malicious). This dynamic generation of clients each epoch, similar to methods used for estimating client relationships \cite{bao2023optimizing}, increases the diversity of simulated scenarios and helps prevent overfitting. Each training iteration within an epoch consists of two main steps: {\it data simulation/attack} and {\it feature extraction/labeling}. 

\paragraph{Step 1: Data Simulation and Attack}
For each sample $x \in \mathcal{D}^{\text{aux}}$, the server decides whether to simulate an attack on that sample or keep it clean. This decision is made by drawing from a Bernoulli distribution with probability $p=1/2$ of being attacked. If selected for attack, the server randomly chooses between two types of data perturbations with equal probability: blur or noise. If a sample is selected for blurring, the server samples a blur severity parameter $\beta \sim \text{Unif}(\beta_0, \beta_1)$ and applies a blurring operation according to Definition \ref{def:blur_attack}. This simulates clients whose data might be of lower quality or intentionally blurred to impair model training or target specific vulnerabilities. If a sample is selected for adding noise, the server samples a noise variance $\sigma \sim \text{Unif}(\sigma_0, \sigma_1)$ and applies additive noise according to Definition \ref{def:noisy_Attack}. This simulates clients whose data might be corrupted by sensor noise or intentionally perturbed with adversarial noise patterns.
After processing all samples in $\mathcal{D}^{\text{aux}}$ in this manner, the server possesses a modified dataset where each sample is either clean, blurred, or noisy, with the attack type and parameters recorded.

\paragraph{Step 2: Fictitious Client Creation and Feature Extraction}
The modified dataset from Step 1 is then partitioned to create the data for $\tilde{K}$ fictitious clients. These clients are equally divided into two groups: $\tilde{K}/2$ benign and $\tilde{K}/2$ malicious. Clean data samples are assigned to benign clients, while attacked data samples (either blurred or noisy) are assigned to malicious clients. Let $\{x_k^i\}_{i=1}^{n_k}$ denote the data points assigned to the $k$-th fictitious client, where $n_k$ is the number of samples for client $k$.

\paragraph{Principal Component Analysis}
For each simulated client \(k\), PCA \cite{abdi2010principal} is applied to their local dataset \(\{x_k^i\}_{i=1}^{n_k}\) to analyze the covariance structure and extract the top \(r\) principal components \(v_k^i\) with eigenvalues \(\lambda_k^i\), capturing dominant directions of variance. A compact representation vector is defined as:
\begin{equation}\label{eq:pca_representative}
   \hat{x}_k = \sum_{i = 1}^{r} \alpha^i_k v_k^i, \quad \text{with} \quad \alpha^i_k = \frac{\lambda_k^i}{\sum_{j=1}^r \lambda_k^j}
\end{equation}
This PCA-derived vector \(\hat{x}_k\) summarizes the client data’s intrinsic structure by weighting principal directions by their explained variance. The PCA step supports dimensionality reduction and noise filtering, extracting features sensitive to structural perturbations such as blur or noise. Notably, it is performed offline on simulated data at the server: in real FL deployments, clients neither share raw data nor PCA results. Instead, this offline PCA informs training, while clients transmit only the privacy-preserving spectral embedding \(\varphi_k\), discussed next.

\paragraph{Spectral Embedding}
Following this PCA step, the spectral representation $\varphi_k$ is computed for each fictitious client $k$. This is achieved by applying a spectral operator $\Phi$ (either the WST or FT) to statistics derived from the client's data distribution, such as the PCA-derived representation vector $\hat{x}_k$ or the set of principal eigenvalues $\lambda_k^i$. Spectral transforms are particularly sensitive to frequency and texture information, making them effective at capturing the systematic changes introduced by attacks like blur and noise. The output $\varphi_k = |\Phi[\hat{x}_k]|$, where the modulus is taken element-wise, results in a fixed-size vector representation for each client. This $\varphi_k$ is designed to be an aggregate statistic that captures characteristics of the data distribution without revealing individual data points, making it suitable as a non-privacy-leaking feature for the detector in a live FL setting.\\
Finally, for each epoch, we obtain a dataset of client representations and their corresponding labels: $\{(\varphi_k, \mu_k)\}_{k=1}^{\tilde{K}}$, where $\mu_k \in \{\text{B (Benign)}, \text{A (Attacker)}\}$. The detector weights $w$ are updated using a stochastic optimizer (e.g., SGD, Adam) to minimize a binary classification loss, such as Binary Cross-Entropy (BCE) \cite{ruby2020binary}, between the detector's prediction based on $\varphi_k$ and the ground-truth label $\mu_k$. 

\begin{algorithm}
\caption{\shortname Offline Training}
\label{alg:waffle_summary}
\begin{algorithmic}[1]
\Require Auxiliary dataset $\mathcal{D}^{\text{aux}}$, Number of epochs $E$, Number of fictitious clients $\tilde{K}$, Number of top PCs $r$, Spectral operator $\Phi$, Learning rate $\eta$
\Ensure Trained detector weights $w$
\vspace{0.2cm}
\State Initialize detector weights $w$ 

\For{$e = 1 \dots E$}
    \State // \textbf{Simulate Data and Clients for Epoch $e$}
    \State $\mathcal{D}^{\text{simulated}}_e \leftarrow \text{SimulateAttackedData}(\mathcal{D}^{\text{aux}})$ \Comment{Applies random attacks to $\mathcal{D}^{\text{aux}}$}
    \State $\{(\mathcal{D}_k, \mu_k)\}_{k=1}^{\tilde{K}} \leftarrow \text{PartitionData}(\mathcal{D}^{\text{simulated}}_e, \tilde{K})$ \Comment{Creates $\tilde{K}$ clients with labels}
\vspace{0.2cm}
    \State // \textbf{Extract Features for Each Simulated Client}
    \State Initialize epoch dataset $\mathcal{S}_e = \emptyset$ \Comment{Stores $(\varphi_k, \mu_k)$ pairs}
    \For{$k = 1 \dots \tilde{K}$}
        \State $\{x_k^i\}_{i=1}^{n_k} \leftarrow \mathcal{D}_k$
        \State Compute PCA-derived representation $\hat{x}_k$ from $\{x_k^i\}$ \Comment{Eq. \eqref{eq:pca_representative}}
        \State Compute spectral embedding $\varphi_k \leftarrow |\Phi[\hat{x}_k]|$ \Comment{Apply FT or WST to $\hat{x}_k$}
        \State Add $(\varphi_k, \mu_k)$ to $\mathcal{S}_e$
    \EndFor
\vspace{0.2cm}
    \State // \textbf{Update Detector Model}
    \State $w\leftarrow \text{Opt}(\mathcal{L}_{\text{BCE}}(w; \mathcal{S}_e))$ \Comment{Optimization step}

\EndFor

\State \Return $w$
\end{algorithmic}
\end{algorithm}
\subsection{Offline Detection and Filtering}\label{sec:detection_phase}
Once the \shortname\ detector model $w$ has been trained offline on the simulated auxiliary dataset $\mathcal{D}^{\text{aux}}$ and prior to the first FL communication round, each client $k \in \{1,\dots,K\}$ in the federation processes its local training data $\{x_k^i\}_{i = 1}^{n_k}$ {\it privately} on their device. This processing involves a sequence of steps performed locally. First, each client computes the PCA of their local training samples to derive the representation vector $\hat{x}_k$, as defined in Equation \eqref{eq:pca_representative}.Then, each client computes its spectral embedding $\varphi_k = \Phi[\hat{x}_k]$, by applying the spectral operator $\Phi$ (WST or FT). \\
After completing these local computations and obtaining $\varphi_k$, each client $k$ securely transmits only this resulting spectral embedding vector to the server. The server, upon receiving $\varphi_k$ from each participating client, inputs it into the pre-trained \shortname detector $w$. Clients that are classified as malicious by the detector are then excluded from participating in the federated training process for the global model $\theta$. This preemptive filtering step enhances the stability and reliability of the global model training process, leading to potentially faster and more robust convergence by ensuring that aggregation occurs over updates from predominantly benign sources.\\ Moreover, due to its modular nature, \shortname operates as an initial defense layer. The set of clients validated as benign by \shortname can  proceed with any federated learning aggregation methods, allowing \shortname\ to be easily combined with other online robust aggregation techniques to further strengthen the overall defense strategy.

\section{Theoretical Guarantees}\label{sec:theoretical_guarantees}

In this section, we establish a theoretical foundation for our proposed algorithm, which we refer to as \shortname. Our primary focus is to demonstrate the benefits of removing adversarial clients in FL scenarios. We show that by filtering out malicious updates, \shortname{} provides a more accurate estimate of the true global model compared to standard \texttt{FedAvg} \cite{mcmahan2017communication}, which is susceptible to adversarial poisoning. We provide general error bounds with detailed proofs presented in Appendix \ref{app:theory}.

Let $\mathcal{B} \subset \{1,\dots,K\}$ denote the set of benign clients and $\mathcal{M} \subset \{1,\dots,K\}$ the set of malicious clients in a federated system with $K$ total clients. We assume these sets are disjoint and their union covers all clients, i.e., $\mathcal{B} \cap \mathcal{M} = \emptyset$ and $\mathcal{B} \cup \mathcal{M} = \{1,\dots,K\}$. To model the heterogeneity and potential adversarial influence in client updates, we adopt the following statistical framework:

\begin{assumption}\label{ass:1}
For each benign client $k \in \mathcal{B}$, the local model update $\theta_k$ is an independent random variable drawn from a distribution $\rho_k(\bar{\theta}^b, \sigma^b)$. This distribution is centered around a common benign mean $\bar{\theta}^b$ with variance $(\sigma^b)^2$, i.e., $\mathbb{E}[\theta_k] = \bar{\theta}^b$ and $\mathbb{V}ar[\theta_k] = (\sigma^b)^2$. Similarly, for malicious clients $k \in \mathcal{M}$, the local updates $\theta_k$ are independent random variables drawn from $\rho_k(\bar{\theta}^m, \sigma^m)$ with $\mathbb{E}[\theta_k] = \bar{\theta}^m$ and $\mathbb{V}ar[\theta_k] = (\sigma^m)^2$.
\end{assumption}

\begin{assumption}\label{ass:2}
We posit that malicious clients exhibit significantly higher update variance compared to benign clients, reflecting a diverse range of attack strategies and the potential for large, destabilizing updates. Formally, we assume $\sigma^m \gg \sigma^b$.
\end{assumption}

The standard federated averaging estimator is defined as a weighted average of client updates: $\theta_{avg} = {1}/{K}\sum_{k = 1}^K  \theta_k$. Our objective is to obtain an estimator that is unbiased with respect to the benign client distribution, meaning $\mathbb{E}[\theta_{avg}] = \bar{\theta}^b$. We demonstrate that removing malicious clients is crucial for achieving this goal. We analyze two scenarios: one where the benign and malicious updates have different means (Lemma \ref{lemma:different_mean}) and one where they share the same mean but differ in variance (Lemma \ref{lemma:same_mean}). 

\begin{lemma}\label{lemma:different_mean}
If the benign and malicious client updates have different mean parameter values, i.e., $\bar{\theta}^m \neq \bar{\theta}^b$, then the standard federated averaging estimator $\theta_{avg}$ is a \textbf{biased estimator} of $\bar{\theta}^b$, meaning $\mathbb{E}[\theta_{avg}] \neq \bar{\theta}^b$.
\end{lemma}

\begin{lemma}\label{lemma:same_mean}
Let $\theta_{avg}^{\mathcal{B}} = \frac{1}{|\mathcal{B}|}\sum_{k \in \mathcal{B}} \theta_k$ be the federated averaging estimator computed using only benign client updates. Under Assumption \ref{ass:2}, if $(\sigma^m)^2 > \left(2 + \frac{|\mathcal{M}|}{|\mathcal{B}|}\right) (\sigma^{b})^2$ , then the variance of the standard federated averaging estimator is higher than that of our estimator:  $\mathbb{V}ar[\theta_{avg}] \geq \mathbb{V}ar[\theta_{avg}^{\mathcal{B}}]$.
\end{lemma}

Lemmas \ref{lemma:different_mean} and \ref{lemma:same_mean} provide the foundation for the following proposition, which formally establishes the advantage of removing malicious clients from the federated aggregation process.

\begin{proposition}\label{prop:removal}
Under Assumptions \ref{ass:1} and \ref{ass:2}, removing malicious clients (those in $\mathcal{M}$) from the federation yields a superior estimator of the global model. Specifically, the resulting estimator is unbiased (in the sense of Lemma \ref{lemma:different_mean}) and exhibits a reduced variance (as shown in Lemma \ref{lemma:same_mean}), leading to improved model accuracy and robustness.
\end{proposition}

\section{Experiments}\label{sec:experiments}
In this section, we present experimental results on widely used federated learning benchmark datasets~\cite{caldas2018leaf, xiao2017fashion, cifar10}, comparing the performance of \shortname in its two variants—one using the WST representation and the other using FT—with established baselines from the Byzantine-resilient FL literature. Details on implementation settings, datasets, and models are provided in Appendix~\ref{app:exp}.\\
Section~\ref{sec:expwstft} evaluates the detection performance of the two variants of \shortname, highlighting the differences between the WST and FT representations. In Section~\ref{sec:expcomparison_baseline}, we compare \shortname against standard Byzantine-resilient FL baselines, including \texttt{FedAvg}~\cite{mcmahan2017communication}, \texttt{Krum} and \texttt{mKrum}~\cite{blanchard2017machine}, \texttt{GeoMed}~\cite{chen2017distributed}, and \texttt{TrimmedMean}~\cite{yin2018byzantine}. Additionally, we demonstrate that \shortname can be applied on top of any aggregation algorithm, improving their performance. Further experiments, comparisons and code release details are reported in Appendix~\ref{app:exp}, and the metrics used for evaluation—both for detection and classification—are detailed in Appendix~\ref{app:metrics}.

\subsection{\shortname : WST vs Fourier}\label{sec:expwstft}
\begin{figure}[h!]
\centering
\includegraphics[width=0.48\linewidth]{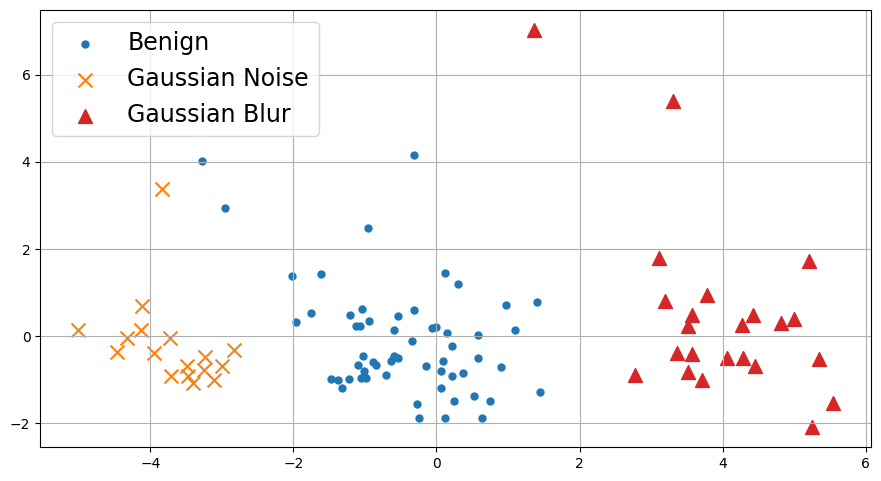}
        \label{fig:le3ft}
    \hfill
\includegraphics[width=0.48\linewidth]{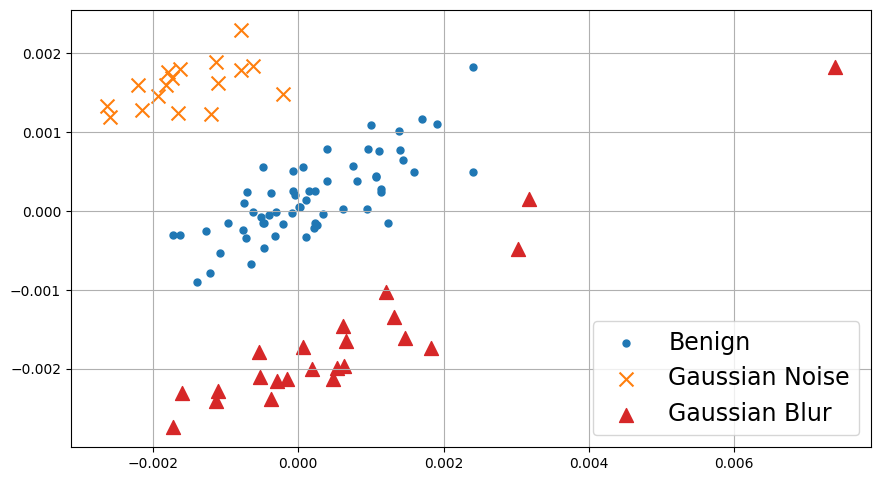}
        \label{fig:rig3ht}
    \caption{\small Client distributions of the  $\varphi_k$ for Cifar10 dataset with $K = 100$ clients on a 2-dimensional space, for \shortname + FT (left), and \shortname + WST (right). There is a total of 60 benign clients (dots), and 40 attackers: 20 noisy (crosses) and 20 blurred (triangles). Both methods provide a noticeable separation between the clients. }
    \label{fig:wst_ft}
\end{figure}

We compare the detection performance of \shortname to assess the differences between the WST and FT representations. As illustrated in Figure~\ref{fig:wst_ft}, both representations yield a clear separation between benign and malicious clients. The visualizations—obtained via two-dimensional PCA embeddings—show that the method effectively distinguishes between the different attacker groups and benign clients, regardless of the chosen representation. However, as shown in Table~\ref{tab:detection_wst}, the quantitative results at the client level differ between the two variants. We report standard detection metrics: precision, F1 score, recall, and accuracy~\cite{lever2016classification}, setting 40\% and 90\% of attackers. The WST variant consistently achieves higher precision and F1 scores, while the FT variant tends to yield higher recall. In the context of malicious client detection, higher recall is often desirable, as it reduces the likelihood of overlooking faulty clients. Table~\ref{tab:detection_wst} highlights the robustness of \shortname: unlike most Byzantine-resilient FL methods, it maintains strong predictive performance even when the vast majority of clients are malicious. Notably, in the extreme case with 90\% adversarial clients, \shortname with WST achieves 100\% precision across all datasets. 
\begin{figure}[h!]
    \centering
    \includegraphics[width=\linewidth]{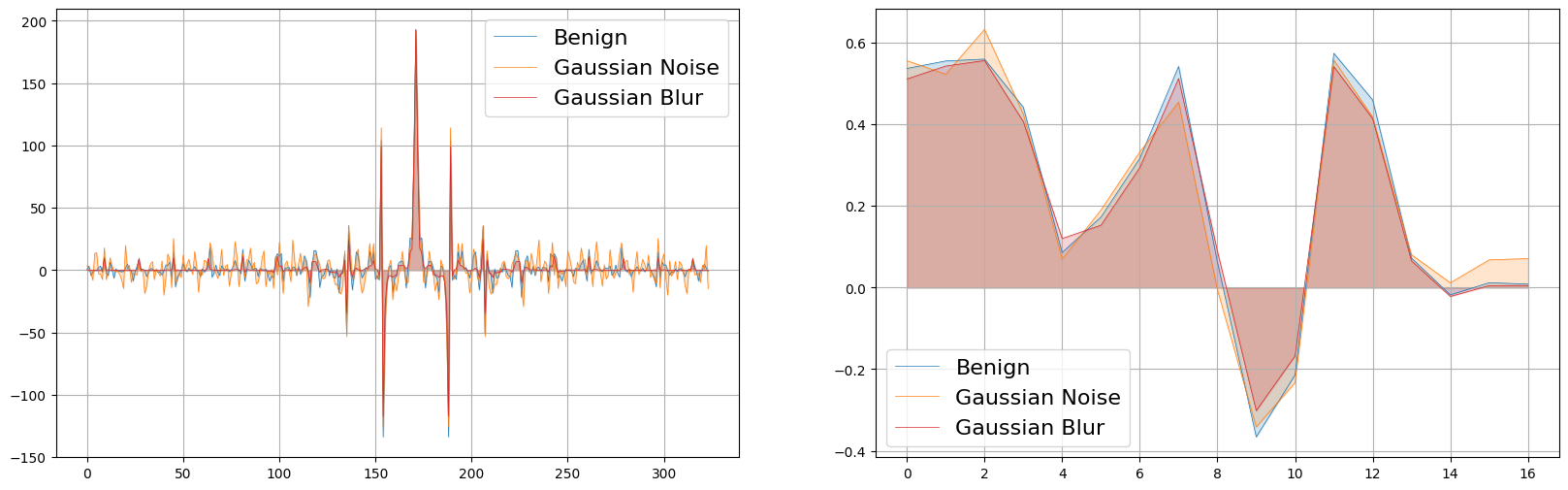}
    \caption{\small Embeddings $\varphi_k$ produced by \shortname for three clients (blur attacker, noise attacker, and benign) on CIFAR-10. The left panel shows the embeddings obtained using FT, while the right panel shows those obtained using WST.}
    \label{fig:eigenfaces}
\end{figure}
\subsection{Comparison with Baselines and Orthogonality of \shortname}\label{sec:expcomparison_baseline}
In this section, we compare \shortname with established Byzantine-resilient FL methods, highlighting its advantages in two complementary settings: (1) we evaluate the impact of applying the two \shortname variants to \texttt{FedAvg}, compared to using different aggregation rules without detection; and (2) we assess the effect of applying \shortname on top of robust aggregation algorithms. As shown in Table~\ref{tab:comparison-baselines-grouped}, the WST variant of \shortname combined with \texttt{FedAvg} consistently outperforms all baselines across all datasets. Furthermore, \shortname improves the performance of each aggregation method it is applied to, demonstrating its orthogonality to the choice of aggregator. These results indicate that \shortname is effective in identifying and removing malicious clients without compromising benign contributions. In contrast, the FT variant exhibits more variable performance, further confirming the suitability of WST representations for this detection task. For reference, we also report the test accuracy of \texttt{FedAvg} trained on a clean federation (i.e., without malicious clients, corresponding to $\theta_{avg}^{\mathcal{B}}$ in the notation of Lemma~\ref{lemma:same_mean}): FashionMNIST 75.08\%, CIFAR-10 50.24\%, CIFAR-100 17.72\%. These values demonstrate that \shortname enables recovery of near-optimal performance, effectively neutralizing the impact of adversarial clients.
\begin{table*}[ht]
\centering
\small
\caption{\small
\textbf{Client Detection.} Comparison between variants of \shortname using WST and FT representations, under two attack scenarios (40\% top, 90\% bottom). Metrics (F1 score, Precision, Recall, Accuracy~\cite{lever2016classification}) refer to the detection of malicious clients. 
}
\label{tab:detection_wst}
\begin{adjustbox}{width=\linewidth}
\begin{tabular}{c|l|cccc|cccc|cccc}
\toprule
& \textbf{Method} 
& \multicolumn{4}{c|}{\textbf{FashionMNIST}} 
& \multicolumn{4}{c|}{\textbf{CIFAR-10}} 
& \multicolumn{4}{c}{\textbf{CIFAR-100}} \\
\cmidrule(lr){3-6} \cmidrule(lr){7-10} \cmidrule(lr){11-14}
& 
& F1 & Prec. & Rec. & Acc. 
& F1 & Prec. & Rec. & Acc. 
& F1 & Prec. & Rec. & Acc. \\
\midrule

\multirow{2}{*}{\rotatebox[origin=c]{90}{\textbf{40\%}}}
& \shortname - FT   & 65.1 \scriptsize{$\pm 3.1$} & 59.9\scriptsize{$\pm 3.1$} & \textbf{69.1} \scriptsize{$\pm 3.1$} & \textbf{69.2}\scriptsize{$\pm 3.1$}  & 80.2\scriptsize{$\pm 2.6$} & 69.1\scriptsize{$\pm 2.6$} & \textbf{96.1}\scriptsize{$\pm 2.6$} & 67.0\scriptsize{$\pm 2.6$} & 55.1 \scriptsize{$\pm 3.2$} & 40.5\scriptsize{$\pm 2.6$} & \textbf{89.7}\scriptsize{$\pm 2.6$} & 44.1\scriptsize{$\pm 2.6$} \\
& \shortname - WST       & \textbf{72.7} \scriptsize{$\pm 1.1$} & \textbf{96.3}\scriptsize{$\pm 1.1$} & 58.2\scriptsize{$\pm 1.1$} & \textbf{82.4}\scriptsize{$\pm 2.6$} & \textbf{95.2}\scriptsize{$\pm 1.0$}  & \textbf{97.6}\scriptsize{$\pm 1.0$} & 92.9\scriptsize{$\pm 1.0$} & \textbf{96.1} \scriptsize{$\pm 1.0$} & \textbf{83.0} \scriptsize{$\pm 1.2$} & \textbf{93.1}\scriptsize{$\pm 1.2$} & 75.1\scriptsize{$\pm 1.2$}  & \textbf{87.0}\scriptsize{$\pm 1.2$}  \\
\midrule

\multirow{2}{*}{\rotatebox[origin=c]{90}{\textbf{90\%}}}
& \shortname - FT    & \textbf{80.9}\scriptsize{$\pm 2.6$} & 94.2 \scriptsize{$\pm 2.6$}& \textbf{70.7}\scriptsize{$\pm 2.6$} &\textbf{71.2}\scriptsize{$\pm 2.6$} & \textbf{93.3}\scriptsize{$\pm 1.6$} & 89.2\scriptsize{$\pm 1.6$} & \textbf{95.7}\scriptsize{$\pm 1.6$}  & \textbf{86.2}\scriptsize{$\pm 1.6$} & 89.0 \scriptsize{$\pm 1.6$}& 88.2\scriptsize{$\pm 1.6$} & \textbf{88.4}\scriptsize{$\pm 1.6$} & \textbf{81.1}\scriptsize{$\pm 1.6$} \\
& \shortname - WST    & 65.6 \scriptsize{$\pm 0.2$} & \textbf{100.0}\scriptsize{$\pm 0.0$} & 49.1\scriptsize{$\pm 0.2$} & 54.0\scriptsize{$\pm 0.2$} & 91.1\scriptsize{$\pm 0.5$} & \textbf{100.0}\scriptsize{$\pm 0.0$} & 83.8 \scriptsize{$\pm 0.5$}& 87.0\scriptsize{$\pm 0.5$} & \textbf{88.1}\scriptsize{$\pm 0.3$} & \textbf{100.0}\scriptsize{$\pm 0.0$} & 68.3\scriptsize{$\pm 0.3$} & 72.2\scriptsize{$\pm 0.3$} \\
\bottomrule
\end{tabular}
\end{adjustbox}
\end{table*}
\begin{table*}
\centering
\caption{\small Comparison between baselines for detecting malicious clients and \shortname (with both WST and FT) with $2\sigma$ error bars. We consider as upper-bound for all methods \texttt{FedAvg} trained on the whole benign federation without malicious clients --- FashionMNIST \textbf{75.5 $\pm$ 1.7}, CIFAR-10 \textbf{50.3 $\pm$ 0.5}, and CIFAR-100 \textbf{17.0 $\pm$ 1.3}.}
\label{tab:comparison-baselines-grouped}
\begin{tabular}{ll|ccccc}
\toprule
\textbf{Dataset} & \textbf{Setting} 
& \texttt{FedAvg} & \texttt{Krum} & \texttt{mKrum} & \texttt{GeoMed} & \texttt{TrimmedMean} \\
\midrule
\multirow{3}{*}{FashionMNIST} 
& w/o detector       & 73.7 $\pm$ 1.3 & 73.8 $\pm$ 1.1 & 72.5 $\pm$ 4.0 & 73.4 $\pm$ 1.7 & 74.6 $\pm$ 0.4 \\
& \shortname - WST   & \textbf{74.9 $\pm$ 1.9} & 70.2 $\pm$ 0.4 & 74.2 $\pm$ 0.9 & 74.6 $\pm$ 1.6 & 74.7 $\pm$ 1.8 \\
& \shortname - FT    & 73.8 $\pm$ 1.1 & 71.4 $\pm$ 2.0 & {74.6 $\pm$ 0.4} & {74.7 $\pm$ 1.0} & \textbf{74.9 $\pm$ 0.5} \\
\midrule
\multirow{3}{*}{CIFAR-10} 
& w/o detector       & 48.7 $\pm$ 1.3 & 44.8 $\pm$ 2.2 & 46.2 $\pm$ 5.9 & 48.3 $\pm$ 0.5 & 48.1 $\pm$ 0.4 \\
& \shortname - WST   & \textbf{49.6 $\pm$ 0.3} & 46.2 $\pm$ 0.6 & {49.5 $\pm$ 0.6} & {49.0 $\pm$ 1.4} & {49.5 $\pm$ 0.8} \\
& \shortname - FT    & 47.1 $\pm$ 0.4 & 43.8 $\pm$ 1.8 & 46.7 $\pm$ 1.3 & 47.2 $\pm$ 0.3 & 46.8 $\pm$ 1.1 \\
\midrule
\multirow{3}{*}{CIFAR-100} 
& w/o detector       & 16.4 $\pm$ 0.1 & 10.1 $\pm$ 0.8 & 14.6 $\pm$ 0.7 & 16.4 $\pm$ 0.7 & 16.5 $\pm$ 1.1 \\
& \shortname - WST   & \textbf{16.5 $\pm$ 1.0} & 8.8 $\pm$ 2.2 & {14.5 $\pm$ 0.7} & 16.3 $\pm$ 0.3 & {16.2 $\pm$ 0.5} \\
& \shortname - FT    & 11.6 $\pm$ 0.2 & {7.6 $\pm$ 0.6} & 10.6 $\pm$ 0.7 & {12.1 $\pm$ 0.3} & 10.6 $\pm$ 0.5 \\
\bottomrule
\end{tabular}
\end{table*}

\subsection{Comparison with Anomaly Detection Baselines}

To provide a comprehensive comparison, we also benchmark our approach against other recent detection methods from the literature, namely \texttt{FLDetector} \cite{zhang2022fldetector}and \texttt{VAEDetector}~\cite{li2020learning}. These methods operate as online techniques, analyzing model updates across multiple training rounds to identify malicious behavior. We evaluate their performance in a scenario with 60\% benign clients under the same random block attack. The accuracy achieved by \texttt{FedAvg} when integrated with these detectors is reported in Table~\ref{tab:detector_comparison}. The results show that under this challenging, non-Gaussian attack scenario, these benchmarks were unable to reliably detect the attacks, leading to a significant drop in performance compared to our method.

\begin{table}[h!]
\centering
\caption{Performance comparison with other detection methods under a random block attack with 40\% malicious clients.}
\label{tab:detector_comparison}
\begin{adjustbox}{width=\linewidth}
\begin{tabular}{lccccc}
\toprule
\textbf{Dataset} & \texttt{FedAvg} & \shortname-WST& \shortname-FT & \texttt{FLDetector} & \texttt{VAEDetector} \\
\midrule
FashionMNIST & $73.7 \pm 1.3$ & $\textbf{74.9} \pm \textbf{1.9}$ & $73.8 \pm 1.1$ & $71.4 \pm 0.9$ & $73.1 \pm 0.8$ \\
CIFAR-10 & $48.7 \pm 1.3$ & $\textbf{49.6} \pm \textbf{0.3}$ & $47.1  \pm 0.4$ & $45.4 \pm 0.6$ & $47.0 \pm 0.5$ \\
CIFAR-100 & $16.4 \pm 0.1$ & $\textbf{16.5} \pm \textbf{1.0}$ & $11.6 \pm 0.2$ & $16.2 \pm 0.1$ & $15.5 \pm 0.7$ \\
\bottomrule
\end{tabular}
\end{adjustbox}

\end{table}

\subsection{\shortname with Non-Gaussian Attacks}
In the theoretical section of the paper, we formalize two scenarios of attacks, i.e., clients providing images that have been perturbed with additive Gaussian noise -- \textit{noisy clients} -- and those that have blurred data, where the blur mathematically is a convolution with a Gaussian kernel. In this section, we provide an experimental evaluation of our method in the presence of non-Gaussian attacks. The experimental framework that we analyze consists of an attack in which a random subset of pixels of each client's data is perturbed; in this case, 50\% are substituted with black pixels.

However, the \texttt{Waffle} framework is not limited to detecting these types of attacks. To validate its robustness against more complex, non-Gaussian structural attacks, we conducted further experiments. In this new scenario, malicious clients apply a random dropout attack on part of the image, where 50\% of the image pixels, grouped into small random blocks, are set to zero. This introduces sharp, non-Gaussian artifacts that are structurally different from simple noise. \texttt{Waffle-WST} obtained an almost perfect detection performance, as it is summarized here.

We report in Table~\ref{tab:block_attack} the results of the random block attack across CIFAR-10, CIFAR-100, and Fashion-MNIST, including standard robust aggregation baselines.

\begin{table*}
\centering
\caption{Performance under random block attack. We report mean test accuracy and 2-sigma error bars over multiple runs. We consider as upper-bound for all methods \texttt{FedAvg} trained on the whole benign federation without malicious clients --- FashionMNIST \textbf{75.5 $\pm$ 1.7}, CIFAR-10 \textbf{50.3 $\pm$ 0.5}, and CIFAR-100 \textbf{17.0 $\pm$ 1.3}.}
\label{tab:block_attack}
\begin{tabular}{lccccccc}
\toprule
\textbf{Dataset} & \texttt{FedAvg}& \texttt{Krum} & \texttt{MultiKrum} & \texttt{TrimmedMean} & \texttt{GeoMed} & \texttt{Waffle-WST} & \texttt{Waffle-FT} \\
\midrule
CIFAR-10 & $48.7 \pm 1.3$ & $44.5 \pm 0.2$ & $47.8 \pm 0.2$ & $47.9 \pm 0.3$ & $48.2 \pm 0.2$ & $\textbf{49.8} \pm\textbf{ 0.2}$ & $48.5 \pm 0.2$ \\
CIFAR-100 & $16.4 \pm 0.1$ & $9.4 \pm 0.1$ & $15.3 \pm 0.1$ & $16.2 \pm 0.1$ & $15.1 \pm 0.1$ & $\textbf{16.9} \pm \textbf{0.1}$ & $16.3 \pm 0.1$ \\
Fashion-MNIST & $73.7 \pm 1.3$ & $74.3 \pm 0.3$ & $71.7 \pm 0.3$ & $75.0 \pm 0.2$ & $71.4 \pm 0.3$ & $\textbf{75.4} \pm \textbf{0.5}$ & $75.3 \pm 0.6$ \\
\bottomrule
\end{tabular}
\end{table*}

As expected from our theoretical results, if the detector is perfect, we reach a performance of the federated training that is comparable with the situation without malicious clients. The \texttt{WST} variant is particularly effective, as it is designed to capture local structural information and textures. The random dropout attack fundamentally disrupts these local patterns, creating a strong and detectable signal for our framework. The benefit of \texttt{Waffle}, especially the WST variant, is particularly prominent on color images (CIFAR-10/100), where the attack disrupts chromatic and texture patterns that \texttt{WST} is well-suited to detect. In contrast, Fashion-MNIST consists of grayscale images, where the attack is more subtle and less disruptive to local statistics. Nonetheless, \texttt{Waffle-WST} still achieves performance very close to the clean-case baseline. This displays that \shortname is a robust solution capable of identifying a broader class of feature-level data integrity attacks beyond simple Gaussian perturbations.
\subsection{\shortname in NLP tasks}
As a proof of concept for tasks beyond computer vision, we extend our evaluation to a Natural Language Processing (NLP) task. We do not compare against other baselines here, as this is intended to demonstrate the versatility of our framework. For this experiment, we implemented a composite Shift-and-Noise Attack on the 50-dimensional GloVe embeddings \cite{pennington2014glove} in its most recent version \cite{carlson2025new} for 40\% of the 100 clients in the federation. The attack consists of two components: (1) applying random permutations to the embedding vectors and (2) adding Gaussian noise. Our \texttt{Waffle-WST} method demonstrated strong detection capabilities against this attack, achieving an F1-score of 0.73 and, notably, a perfect precision of 1.0, ensuring no honest clients were penalized. This successful detection directly translated to a significant performance recovery in the global model. As shown in Table~\ref{tab:nlp_results}, the \texttt{Waffle} detector is able to raise the final test accuracy from a compromised 38.53\% (without our detector) to 42.71\%. This result brings the model's performance remarkably close to the ideal, attack-free scenario of 44.81\%. These outcomes are entirely consistent with the extensive experiments in our main paper, and we will provide a full and detailed analysis of these NLP results in the camera-ready submission. \begin{table}[h!] \centering \caption{Model accuracy on the NLP task under a composite Shift-and-Noise attack. The task is classification of sentiments, therefore the evaluation metric is still accuracy.} \label{tab:nlp_results} \begin{tabular}{lc} \toprule \textbf{Scenario and method} & \textbf{Test Accuracy (\%)} \\ \midrule \texttt{FedAvg} w/o malicious clients (No Attack) & $44.81 \pm 2.1 $ \\ \texttt{FedAvg} w/ \shortname-WST (Under Attack) & $42.71 \pm 1.9$ \\ \texttt{FedAvg} w/ \shortname-FT (Under Attack) & $41.91 \pm 1.6$ \\ \texttt{FedAvg} w/o Detector (Under Attack) & $38.53 \pm 2.0$ \\ \bottomrule \end{tabular} \end{table}

\section{Conclusion}
We propose \shortname, a novel offline algorithm to detect malicious client data in FL before training commences. Exploiting stable spectral features extracted via the WST and FT, our method enables robust anomaly detection from private, low-dimensional client-side summaries built on publicly distilled data. By filtering out compromised clients prior to the aggregation process, \shortname significantly improves convergence speed, final model accuracy, and robustness to data contamination. Our benchmarks show it achieves near-perfect precision, even in extreme scenarios with up to 90\% malicious clients, outperforming strategies that rely solely on robust aggregation.

A key advantage of \shortname is its role as a proactive and complementary security layer. By specializing in the detection of data-level attacks, it acts as an essential first line of defense, sanitizing the client pool before resource-intensive training begins. This model-agnostic approach is not intended to replace in-training defenses but rather to fortify them. It can be seamlessly integrated with existing FL defenses, such as robust aggregation mechanisms that target model-level threats, to create a more comprehensive, multi-layered security architecture against a wider spectrum of attacks.

This early-detection mechanism also yields substantial practical benefits by reducing training time, communication overhead, and energy consumption—factors that are crucial in large-scale and resource-constrained deployments, like IoT. By enhancing the robustness, trustworthiness, and efficiency of the training pipeline, our method helps pave the way for secure FL deployments in sensitive domains like connected healthcare, autonomous systems, and smart infrastructure.

Future work will focus on extending \shortname to defend against more sophisticated threats, including backdoor attacks, model poisoning, and sybil-based infiltration. In parallel, we plan to adapt the approach to support wider neural architectures capable of handling more complex and high-dimensional datasets, such as CIFAR-100 or even ImageNet-scale benchmarks. These directions aim to broaden the applicability and impact of \shortname in advancing secure and efficient decentralized machine learning.

\section*{Acknowledgements}
A.L. and D.C. worked under the auspices of Italian National Group of Mathematical Physics (GNFM) of INdAM. A.L. was supported by the Project Piano Nazionale di Ripresa e Resilienza - Next Generation EU (PNRR-NGEU) from Italian Ministry of University and Research (MUR) under Grant DM 117/2023. D.C. expresses their gratitude to Marylou Gabrié for the support.
\bibliography{IEEE-Transactions-LaTeX2e-templates-and-instructions/references}
\bibliographystyle{ieeetr}
\newpage
\appendix
\setcounter{lemma}{0}
\renewcommand{\thelemma}{A\arabic{lemma}}

\setcounter{assumption}{0}
\renewcommand{\theassumption}{A\arabic{assumption}} 

\setcounter{proposition}{0}
\renewcommand{\theproposition}{A\arabic{proposition}} 

\subsection{Theoretical Guarantes}\label{app:theory}
In this Appendix, we present comprehensive proofs and assumptions pertaining to the results delineated in Section \ref{sec:theoretical_guarantees}. Specifically, we focus on demonstrating that in the context of the federated averaging estimator, the exclusion of malicious clients prior to the training process facilitates the attainment of an unbiased estimator of the true mean $\theta^b$, consequently resulting in a less noisy estimate, as stated in Proposition \ref{prop:removal}. This outcome is supported through the introduction of two lemmas, namely Lemmas \ref{lemma:different_mean} and \ref{lemma:same_mean}. 
Without loss of generality, we propose two general assumptions, which reflect the distinct behaviors of benign and malicious clients concerning model distribution. Specifically, we denote the set of benign clients as $\mathcal{B} \subset \{1,\dots,K\}$ and the set of malicious clients as $\mathcal{M} \subset \{1,\dots,K\}$, within a federated system comprising $K$ clients in total. We assume these sets are mutually exclusive and collectively exhaustive, meaning $\mathcal{B} \cap \mathcal{M} = \emptyset$ and $\mathcal{B} \cup \mathcal{M} = \{1,\dots,K\}$. To adequately address the heterogeneity and the potential adversarial impact on client updates, we employ the following statistical framework. 
\begin{assumption}\label{app:ass:1}
For each benign client $k \in \mathcal{B}$, the local model update $\theta_k$ is an independent random variable drawn from a distribution $\rho_k(\bar{\theta}^b, \sigma^b)$. This distribution is centered around a common benign mean $\bar{\theta}^b$ with variance $(\sigma^b)^2$, i.e., $\mathbb{E}[\theta_k] = \bar{\theta}^b$ and $\mathbb{V}ar[\theta_k] = (\sigma^b)^2$. Similarly, for malicious clients $k \in \mathcal{M}$, the local updates $\theta_k$ are independent random variables drawn from $\rho_k(\bar{\theta}^m, \sigma^m)$ with $\mathbb{E}[\theta_k] = \bar{\theta}^m$ and $\mathbb{V}ar[\theta_k] = (\sigma^m)^2$.
\end{assumption}

\begin{assumption}\label{app:ass:2}
We posit that malicious clients exhibit significantly higher update variance compared to benign clients, reflecting a diverse range of attack strategies and the potential for large, destabilizing updates. Formally, we assume that there exists $C>0$ such that $(\sigma^m)^2 > C (\sigma^{b})^2> \left(2 + \frac{|\mathcal{M}|}{|\mathcal{B}|}\right) (\sigma^{b})^2$.
\end{assumption}
Assumption \ref{app:ass:1} characterizes the statistical distributions corresponding to the two distinct groups of clients. In contrast, Assumption \ref{app:ass:2} provides that the variance associated with the malicious models is significantly greater than that of the benign client updates. The standard federated averaging estimator is defined as a weighted average of client updates, i.e.
\begin{equation}\label{eq:theta_avg}
    \theta_{avg} = \dfrac{1}{K}\sum_{k = 1}^K  \theta_k \quad .
\end{equation}
Our objective is to obtain an estimator that is unbiased with respect to the benign client distribution, meaning $\mathbb{E}[\theta_{avg}] = \bar{\theta}^b$. We demonstrate that removing malicious clients is crucial for achieving this goal. We analyze two scenarios: one where the benign and malicious updates have different means (Lemma \ref{app:lemma:different_mean}) and one where they share the same mean but differ in variance (Lemma \ref{app:lemma:same_mean}). 
\begin{lemma}\label{app:lemma:different_mean}
If the benign and malicious client updates have different mean parameter values, i.e., $\bar{\theta}^m \neq \bar{\theta}^b$, then the standard federated averaging estimator $\theta_{avg}$ is a \textbf{biased estimator} of $\bar{\theta}^b$, meaning $\mathbb{E}[\theta_{avg}] \neq \bar{\theta}^b$.
\end{lemma}
\begin{proof}
Let us first recall that a random variable $\hat{X}$ is an unbiased estimator of $\mu$, if its expectation equals the parameter that we aim to estimate, i.e. if $\mathbb{E}[\hat{X}] = \mu$. In case $\mathbb{E}[\hat{X}] \neq \mu$, we say that $\hat{X}$ is a biased estimator of $\mu$.

If we compute the expectation of the estimator $\theta_{avg}$, defined in Equation \ref{eq:theta_avg}, using the fact that malicious and benign client $\{\mathcal{B},\mathcal{M}\}$ form a partition of $\{1,\dots,K\}$, we get
\begin{equation}
    \mathbb{E}[\theta_{avg}] = \mathbb{E}\left[\dfrac{1}{K} \sum_{k = 1}^K \theta_k\right] = \mathbb{E}\left[\dfrac{1}{K}\left( \sum_{k \in \mathcal{B}} \theta_k +  \sum_{k \in \mathcal{M}} \theta_k\right)\right]\quad.
\end{equation}
Let us denote with $M = |\mathcal{M}|$ and $B = |\mathcal{B}|$ the number of malicious and benign clients in the federation, by exploiting linearity of the expectation operator, we obtain
\begin{equation}
\begin{split}
    \mathbb{E}[\theta_{avg}] &= \dfrac{1}{K}\left(\sum_{k \in \mathcal{B}} \mathbb{E}[\theta_k] + \sum_{k \in \mathcal{M}} \mathbb{E}[\theta_k]\right) = \dfrac{B \bar{\theta}^b + M \bar{\theta}^m}{K} \\
    &= \dfrac{B \bar{\theta}^b + M \bar{\theta}^b - M \bar{\theta}^b + M \bar{\theta}^m}{K} \\ &= \bar{\theta}^b + \dfrac{M}{K}(\bar{\theta}^m - \bar{\theta}^b) \neq \bar{\theta}^b
\end{split}
\end{equation}
where $\bar{\theta}^b$ and $\bar{\theta}^m$ denote the expectation of the model updates for benign and malicious clients, respectively. Since we obtained that $\mathbb{E}[\theta_{avg}] \neq \bar{\theta}^b$, we conclude that the estimator is biased.
\end{proof}
Furthermore, we observe that the drift in the estimate away from the benign model is controlled by the ratio of malicious clients $M$ with respect to the number of total clients $K$.

\begin{lemma}\label{app:lemma:same_mean}
Let
\begin{equation}
    \theta_{avg}^{\mathcal{B}} = \frac{1}{|\mathcal{B}|}\sum_{k \in \mathcal{B}} \theta_k
\end{equation}
be the federated averaging estimator computed using only benign client updates. Under Assumption \ref{ass:2} the variance of the standard federated averaging estimator is higher than that of our estimator:  $\mathbb{V}ar[\theta_{avg}] \geq \mathbb{V}ar[\theta_{avg}^{\mathcal{B}}]$.
\end{lemma}
\begin{proof}
First, we compute the variance for the two estimators $\theta_{avg}$ and $\theta_{avg}^{\mathcal{B}}$ exploiting the indipendece between model distributions, posit in Assumption \ref{app:ass:1}. In particular
\begin{equation}
\begin{split}
    \var[\theta_{avg}] &= \var\left[\dfrac{1}{K}\sum_{k = 1}^K \theta_k\right]  \\ &=\dfrac{1}{K^2}\left( \sum_{k \in \mathcal{B}}\var[\theta_k] + \sum_{k \in \mathcal{M}}\var[\theta_k] \right) \\&= \dfrac{B (\sigma^b)^2 + M (\sigma^m)^2}{K^2}\quad.
\end{split}
\end{equation}
Similarly, we get that
\begin{equation}
    \var[\theta_{avg}^\mathcal{B}] = \dfrac{(\sigma^b)^2}{B}\quad. 
\end{equation}
If we consider the difference between the variances $\var[\theta_{avg}]$ and $\var[\theta_{avg}^\mathcal{B}]$, and we impose that this quantity is positive qwe obtain the following inequality
\begin{equation}
    \var[\theta_{avg}] - \var[\theta_{avg}^\mathcal{B}] = \dfrac{B (\sigma^b)^2 + M (\sigma^m)^2}{K^2} - \dfrac{(\sigma^b)^2}{B} > 0\quad.
\end{equation}
Recalling that $K = B+ M$, we get
\begin{equation}
\begin{split}
    \dfrac{B^2 (\sigma^b)^2 + M B (\sigma^m)^2 - (B + M)^2 (\sigma^b)^2}{B(B + M)^2} &> 0   \\\iff M B (\sigma^m)^2 - M(2B + M)(\sigma^b)^2 &> 0
\end{split}
\end{equation}
that is, since $M > 0$,
\begin{equation}
    (\sigma^m)^2 > \dfrac{1}{B}(2B + M) (\sigma^b)^2 \iff (\sigma^m)^2 > \left(2 + \dfrac{M}{B}\right)(\sigma^b)^2\quad 
\end{equation}
which together with Assumption \ref{app:ass:1} concludes the proof.
\end{proof}
Lemma \ref{app:lemma:same_mean} provides a definitive bound on the variance of the model, thereby resolving the question \textit{how much larger should the variance of malicious models should be with respect to the benign models' variance.} Nonetheless, given the assumption in \ref{app:ass:2} that the variance of malicious models $\sigma^m$ may exceed that of benign models $\sigma^b$ to an arbitrary extent, the hypothesis of Lemma \ref{app:lemma:same_mean} proves to be non-restrictive and readily achievable.
\begin{proposition}\label{app:prop:removal}
Under Assumptions \ref{ass:1} and \ref{ass:2}, removing malicious clients (those in $\mathcal{M}$) from the federation yields a superior estimator of the global model. Specifically, the resulting estimator is unbiased (in the sense of Lemma \ref{lemma:different_mean}) and exhibits a reduced variance (as shown in Lemma \ref{lemma:same_mean}), leading to improved model accuracy and robustness.
\end{proposition}
\begin{proof}
The proof is immediately derived from Lemmas \ref{app:lemma:different_mean} and \ref{app:lemma:same_mean}. This is due to the fact that upon the exclusion of malicious clients, the federated averaging estimator reduces to the form presented in $\theta_{avg}^{\mathcal{B}}$, which is not only unbiased but also exhibits reduced variance—thereby diminishing noise in the global model's estimation.
\end{proof}

\subsection{Datasets and Implementation Details}\label{app:exp}
We conducted experiments on common FL benchmark datasets \cite{caldas2018leaf}, namely FashionMNIST \cite{xiao2017fashion}, CIFAR-10, and CIFAR-100 \cite{cifar10}. Since our goal was to detect malicious clients based on data characteristics, we sampled benign clients using a Dirichlet distribution with parameter $\alpha = 1000$ to ensure near-i.i.d. conditions. This setting ensures a fair comparison with the baselines, as \shortname neither relies on model updates nor is affected by class imbalance. Moreover, the shuffled training on a distillated dataset already exposes \shortname to synthetic heterogeneity. Introducing additional data imbalance would therefore not yield further insights into its performance.

For classification, we used LeNet-5 \cite{lecun1995comparison}. The standard version was applied to FashionMNIST, while we adjusted the input channels to three and modified the number of output classes for CIFAR-10 and CIFAR-100. The federation included $K = 100$ clients, with $|\mathcal{P}_t| = 10$ clients participating per round. Training was carried out over $T = 500$ communication rounds, using $S = 1$ local epoch per round and a batch size of 64. We employed the cross-entropy loss optimized with the ADAM optimizer \cite{kingma2014adam}, using an initial learning rate $\eta = 0.001$.

For \shortname, we used WST parameters $J = 3$, $L = 6$, and first-order coefficients \cite{kymatio}. The FT baseline employed a window size of 0.5. The \shortname detector was a multilayer perceptron with three hidden layers and hyperbolic tangent activations, trained for 100 epochs using ADAM. The attack parameters $\beta$ and $\sigma$ were randomly sampled from $\text{Unif}\{3, 5, \dots, 19\}$ and $\text{Unif}(0.5, 2.0)$, respectively. For baselines, we used \texttt{mKrum} with $k = 5$, and \texttt{TrimmedMean} with a cut-off parameter of 0.2.

All computations were performed using the CPU of a MacBook Pro equipped with an Apple M3 Pro chip. No additional computational resources were employed.

\subsection{Detection Metrics}\label{app:metrics}

In evaluating the performance of \shortname, we employed several detection metrics \cite{hossin2015review, lever2016classification}, each offering a different perspective on the detector's effectiveness in binary classification task. Let TP, TN, FP, and FN represent True Positives (correctly identified malicious clients), True Negatives (correctly identified benign clients), False Positives (benign clients incorrectly flagged as malicious), and False Negatives (malicious clients incorrectly flagged as benign), respectively.

\paragraph{Accuracy}
Accuracy is one of the most straightforward metrics, representing the overall correctness of the classifier. It is calculated as the ratio of correctly classified instances (both malicious and benign) to the total number of instances.
$$ \text{Accuracy} = \frac{\text{TP} + \text{TN}}{\text{TP} + \text{TN} + \text{FP} + \text{FN}} $$
While intuitive, accuracy can be misleading, especially in scenarios with imbalanced datasets. For instance, if 90\% of clients are benign, a detector that classifies all clients as benign would achieve 90\% accuracy, despite failing to identify any malicious clients. Therefore, while providing a general overview, accuracy alone is often insufficient for evaluating a malicious client detector.

\paragraph{Precision}
Precision measures the proportion of correctly identified malicious clients among all clients classified as malicious by the detector.
$$ \text{Precision} = \frac{\text{TP}}{\text{TP} + \text{FP}} $$
High precision indicates a low false positive rate, meaning that when the detector flags a client as malicious, it is highly likely to be correct. This is crucial in scenarios where incorrectly blocking a benign client (a false positive) has significant negative consequences, such as denying service to legitimate users. A low precision score suggests the detector raises many false alarms.

\paragraph{Recall}
Recall measures the proportion of actual malicious clients that are correctly identified by the detector.
$$ \text{Recall} = \frac{\text{TP}}{\text{TP} + \text{FN}} $$
High recall indicates a low false negative rate, meaning the detector successfully identifies a large fraction of the malicious clients present. This is critical in security applications where failing to detect a malicious client (a false negative) can lead to significant damage or compromise. A low recall score suggests the detector misses many malicious clients.

\paragraph{F1-Score}
The F1-Score is the harmonic mean of Precision and Recall, providing a single metric that balances both concerns.
$$ \text{F1-Score} = 2 \times \frac{\text{Precision} \times \text{Recall}}{\text{Precision} + \text{Recall}} = \frac{2 \times \text{TP}}{2 \times \text{TP} + \text{FP} + \text{FN}} $$
The F1-Score is particularly useful when there is an uneven class distribution, as it punishes extreme values of precision or recall. A high F1-Score indicates that the detector has both good precision and good recall, meaning it is both accurate in its positive predictions and captures a majority of the actual positive instances. It is often preferred over accuracy in imbalanced malicious client detection scenarios where both minimizing false alarms and maximizing detection of actual threats are important.

\subsection{Privacy of \shortname}\label{app:privacy}
\shortname detector architecture prioritizes client privacy throughout its operation. Throughout the learning process, \textbf{individual raw data $\{x_k^i\}$ remains strictly on the client's device}. Each client $k$ \textit{privately} computes its PCA-derived representation $\hat{x}_k$ and subsequently its spectral embedding $\varphi_k$ locally on its own hardware. Clients only transmit the resulting spectral embedding vector $\varphi_k$ to the server, ideally over a secure communication channel to protect these embeddings while in transit. This $\varphi_k$ is explicitly designed to be an aggregate statistic that captures characteristics of the data distribution without revealing individual data points, thereby serving as a non-privacy-leaking feature. Furthermore, since WST is non-invertible, it is also impossible to reconstruct the PCA representant. Our methodology is consistent with approaches in privacy-preserving machine learning where transformed or aggregated representations of data are used instead of raw sensitive information to train models or make inferences \cite{bonawitz2017practical}. Furthermore, the offline training of the \shortname detector and Algorithm~\ref{alg:waffle_summary}) is conducted on a distinct auxiliary dataset $\mathcal{D}^{\text{aux}}$, coherently with common practices \cite{wang2018dataset}, ensuring that no actual client data from the federation is used or exposed during the detector's training phase. The combination of local feature extraction by clients, the transmission of only these specialized spectral embeddings, and offline training using auxiliary data ensures that \shortname functions as a privacy-conscious safeguard within the FL ecosystem.

\end{document}